\documentclass[10pt,twocolumn,letterpaper]{article}

\PassOptionsToPackage{table}{xcolor}
\usepackage[pagenumbers]{cvpr}
\usepackage{multirow}
\usepackage{bbding}
\usepackage{microtype}

\ExplSyntaxOn
\NewDocumentCommand{\AppendPDFPages}{m}
{
	\onecolumn
	\pdfximage{#1}
	\int_step_inline:nn {\pdflastximagepages}
	{
		\AddToShipoutPicture*{
			\AtPageLowerLeft{\includegraphics[page=##1]{#1}}
		}
		\thispagestyle{empty}
		\null\clearpage
	}
}
\ExplSyntaxOff

\definecolor{cvprblue}{rgb}{0.21,0.49,0.74}
\usepackage[pagebackref,breaklinks,colorlinks,allcolors=cvprblue]{hyperref}

\graphicspath{{figures/}}
\def\confName{CVPR}
\def\confYear{2027}

\definecolor{deepred}{RGB}{150,0,0}
\definecolor{deepgreen}{RGB}{0,150,0}
\newcommand{\deepred}[1]{\textcolor{deepred}{#1}}
\newcommand{\deepgreen}[1]{\textcolor{deepgreen}{#1}}

\title{DACA: Dual-Anchor Collaborative Agent for Zero-Shot 3D Visual Grounding}

\author{Jiaming Liu$^{1,3}$\quad Linghe Kong$^{1,\ast}$\quad Guixu Zhang$^{2}$\quad Huaxi Huang$^{3}$\\
	$^1$Shanghai Jiao Tong University\quad $^2$East China Normal University\quad \\$^3$Shanghai Artificial Intelligence Laboratory\\
	\texttt{\{jmliu99,linghe.kong\}@sjtu.edu.cn,huanghuaxi@pjlab.org.cn}
}

\begin{document}
\maketitle
	
\begin{abstract}
	3D visual grounding (3D-VG) is a task that bridges the natural language and 3D space, serving as a foundational capability for embodied intelligence and human-computer interaction. Although recent 3D-VG agents overcome the reliance on pre-processed 3D point clouds, they essentially employ a unidirectional, pipeline-style tool-calling architecture. This architecture separates semantic understanding from geometric reconstruction, leading to error accumulation and an inability to autonomously recover from failures. To address this flaw, this paper proposes the \textbf{D}ual-\textbf{A}nchor \textbf{C}ollaborative \textbf{A}gent (DACA), which reinterprets 3D-VG as a semantic-geometric dual-anchor collaborative reasoning paradigm. DACA comprises two complementary specialized tools: the semantic anchor tool holds the visual semantic features of the target and achieves efficient target tracking via optical flow and VLM local verification; the geometric anchor tool holds the 3D spatial geometric features of the target and performs global spatial reasoning via voxelization and multi-keypoint projection. The two components exchange information bidirectionally via a standardized tool call protocol, autonomously deciding on the next perception-action step to form a closed-loop optimization. Experiments on the ScanRefer and Nr3D benchmarks demonstrate that zero-shot DACA achieves an overall Acc@0.25 of 73.6\% on ScanRefer and 70.4\% overall accuracy on Nr3D, far surpassing the latest zero-shot 3D-VG methods. The code and models are available at \url{https://github.com/liujia99/DACA}.
\end{abstract}

\section{Introduction}
3D visual grounding (3D-VG) links a natural-language description to its referent in 3D space \cite{chen2020scanrefer,achlioptas2020referit3d}. In practical applications such as embodied intelligent navigation \cite{liu2025revisiting}, human-robot collaborative interaction \cite{chen2025synergai}, and augmented reality experiences \cite{arslan2025examining}, intelligent systems must precisely localize and manipulate objects in complex, dynamic 3D environments based on natural language instructions provided by humans. Much of 3D-VG research has relied on supervised methods trained on large-scale annotated data \cite{wu2023eda,zhu2024scanreason,zhu2025llava}. However, such methods are highly dependent on large-scale, high-quality 3D-text aligned annotated data, and the cost of annotating 3D data is 5–10 times that of 2D data \cite{dai2017scannet,wu2023self,liu2023inter,liu2025improving}, which fundamentally limits the generalization performance of supervised methods in environments with open vocabularies.
\begin{figure*}[t]
	\centering
	\includegraphics[width=\linewidth]{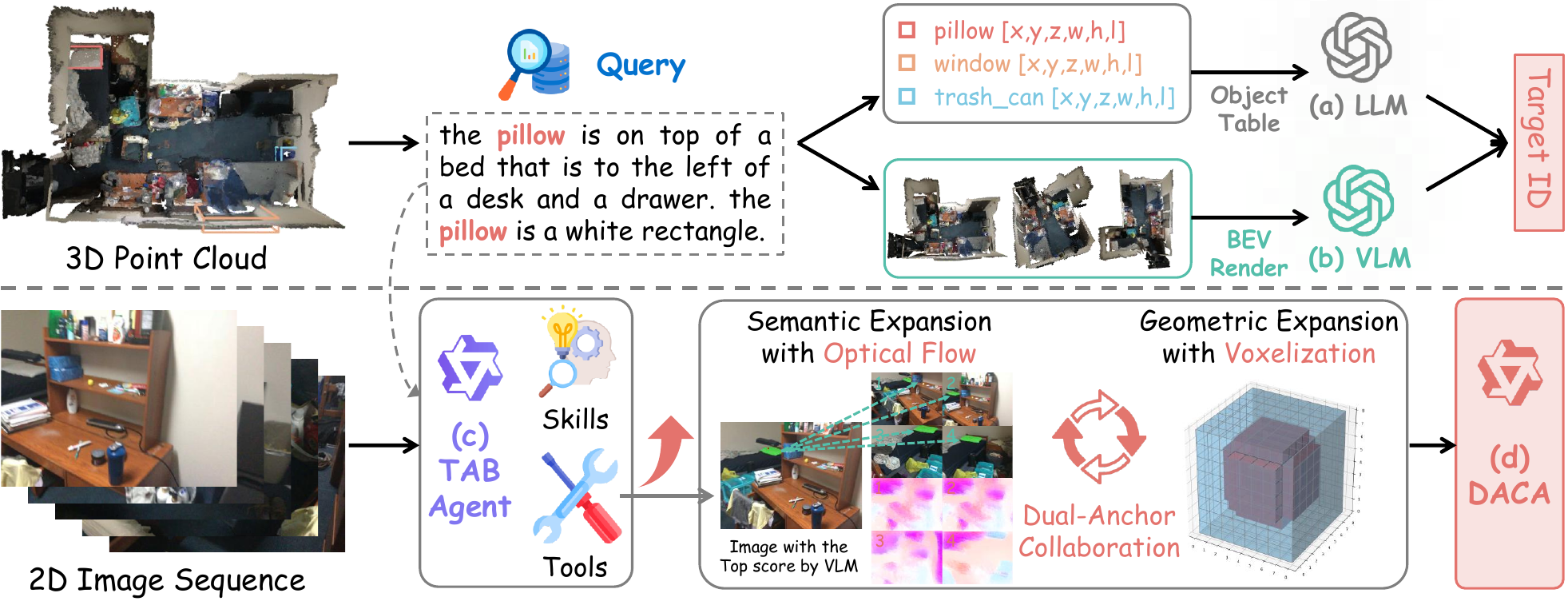}
	\caption{Comparison of four zero-shot 3D-VG paradigms. \textbf{(a) LLM} matches a query to 3D candidate descriptions, and \textbf{(b) VLM} uses rendered multi-view BEV images. \textbf{(c) TAB} passes semantic evidence to geometric extension through the coarse centroid. \textbf{(d) DACA} also makes geometrically retrieved views available for subsequent semantic acquisition. The arrows describe evidence flow between anchors.}
	\label{fig1}
\end{figure*}

To overcome the bottleneck of scarce 3D annotated data, zero-shot 3D-VG has emerged as a hot research topic in recent years \cite{yang2024llm,yuan2024visual,zhang2024agent3d,xu2025vlm}. These methods leverage the powerful generalization capabilities of large language models (LLMs) and visual-language models (VLMs) to achieve object grounding without the need for training on specific scenarios. TAB \cite{wang2026think} formulates 3D-VG as active semantic reasoning and geometric reconstruction instead of static candidate-box matching. By directly processing raw RGB-D streams, it eliminates the reliance on preprocessed 3D point clouds. The proposed semantic-anchored geometric extension mechanism effectively addresses the issue of insufficient multi-view coverage in pure semantic tracking by projecting the 3D center of mass obtained from semantic tracking onto unobserved frames, achieving advanced performance among zero-shot methods on the ScanRefer \cite{chen2020scanrefer} and Nr3D \cite{achlioptas2020referit3d} benchmarks.

However, although TAB has achieved remarkable progress, our in-depth analysis reveals three inherent limitations that restrict its applicability to real-world scenarios:

\noindent\textbf{1. Error accumulation}: TAB's anchor evidence flows from semantic temporal extension to centroid extraction and multi-view geometric extension, although the controller can retry calls or acquire a new reference. A wrong semantic association can bias the centroid and subsequent geometric retrieval. The one-way anchor transfer does not explicitly feed retrieved views back to correct that track.

\noindent\textbf{2. Assumption limitations}: TAB assumes that objects remain stationary within a sequence; its semantic temporal extension relies on frame-by-frame semantic consistency and does not model visual changes between the target and the camera caused by relative viewpoints. Changes in viewpoints may disrupt inter-frame semantic consistency, causing verification to fail and terminating tracking.

\noindent\textbf{3. Projection vulnerability}: TAB’s multi-view geometric extension relies entirely on a single 3D centroid for projection, ignoring spatial uncertainties caused by segmentation errors and object self-occlusion. Errors in the centroid can lead to inaccurate projections, especially under partial occlusion or noisy depth.

In light of the above analysis, this paper proposes the Dual-Anchor Collaborative Agent (DACA) framework, which upgrades zero-shot 3D-VG from a static serial paradigm to a dynamic interactive paradigm, as illustrated in Figure \ref{fig1}. Our core insight is that semantic understanding and geometric reconstruction are not a unidirectional dependency but rather a bidirectional process in which the two components mutually reinforce and correct each other. Specifically, DACA introduces a dual-anchor collaborative mechanism comprising semantic anchors and geometric anchors: semantic anchors capture the visual regions of interest (ROI) of a target and expand via VLM and optical flow, while geometric anchors model the target’s multi-view 3D spatial positions by multi-keypoint projection and voxelization. The agent can revisit acquisition with these observations, which creates a route to correct some errors propagated from the initial track. MG-STE and KA-MGE adapt existing optical-flow and geometric operators. The distinguishing hypothesis is that feeding new geometric observations back into semantic acquisition improves grounding.

In summary, the contributions of this paper are as follows:
\begin{itemize}[leftmargin=*,topsep=0pt, itemsep=0pt]
	\item This work proposes the DACA framework, which introduces a semantic-geometric bidirectional collaborative agent in zero-shot 3D-VG, addressing the issue of unidirectional error accumulation.
	\item Motion-guided semantic temporal extension (MG-STE): Dense optical flow-driven semantic propagation is integrated with VLM to achieve continuous object grounding in 2D sequence consisting of images that cover a scene.
	\item Keypoint-aware multiview geometric extension (KA-MGE): Multi-point geometric representation is constructed with the corner points and centroid of the target’s 3D bounding box, which combined with volumetric scene modeling to achieve precise visibility determination.
	\item Zero-shot evaluation: The open-source model-based DACA outperforms existing zero-shot methods and surpasses fully supervised baselines in grounding 3D objects.
\end{itemize}

\begin{figure*}[t]
	\centering
	\includegraphics[width=\linewidth]{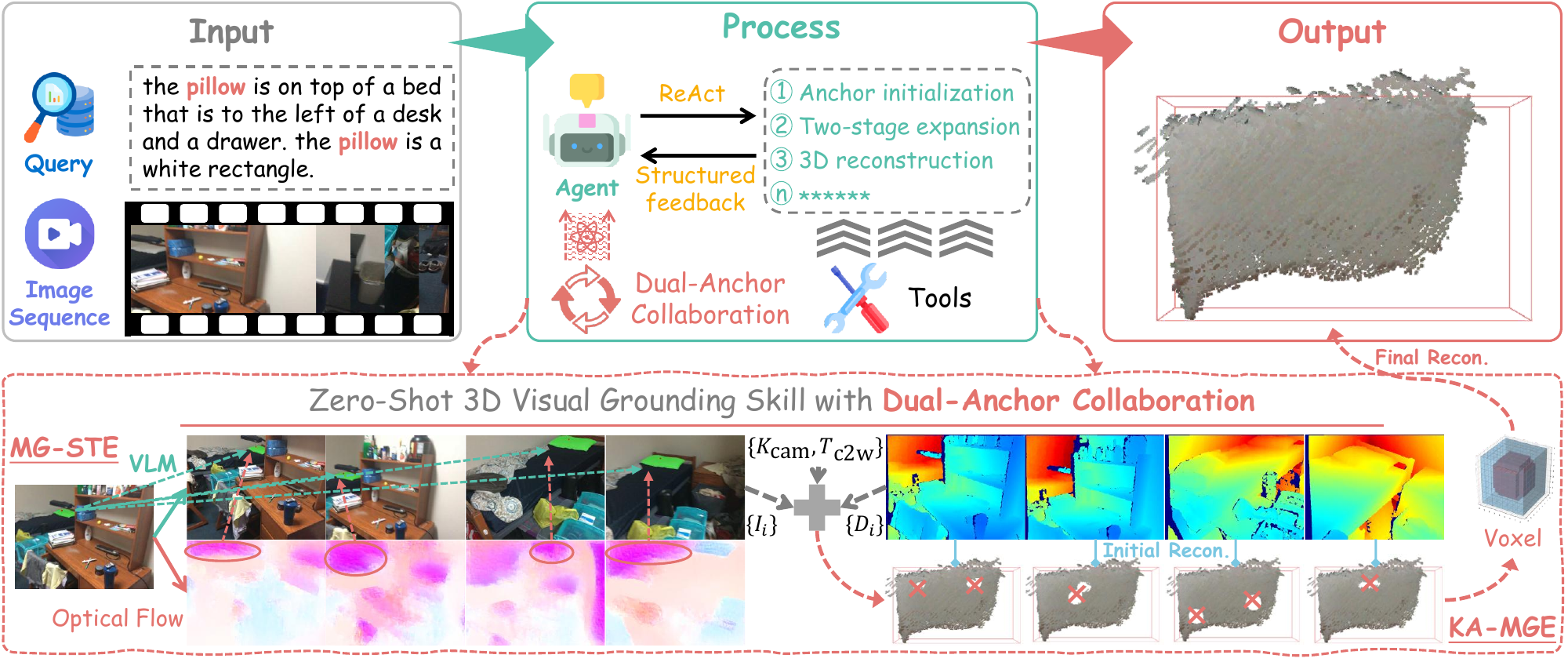}
	\caption{The pipeline of DACA. Given a language query and a sequence of images as input, the agent reasons about 3D-VG tasks under the guidance of 3D-VG skills, utilizing tools such as semantic and geometric anchors to complete the 3D reconstruction and grounding task.}
	\vspace{-10pt}
	\label{fig2}
\end{figure*}

\section{Related Work}
\noindent\textbf{3D visual grounding.} Based on the supervision setting, 3D-VG can be categorized into two major types: fully supervised and zero-shot. Fully supervised methods have been the mainstream in early research and primarily include two-stage and single-stage architectures: two-stage methods \cite{chen2022d,jain2022bottom} first generate candidate regions using 3D detectors and then perform cross-modal feature matching; single-stage methods \cite{wu2023eda,qian2024multi} directly fuse point cloud and text features to output the target bounding box. Although these methods achieve high grounding accuracy on closed datasets, their generalization ability is strictly limited by the coverage of the training data, making it difficult to transfer to unseen object categories and scenes. In contrast, the core objective of zero-shot 3D-VG is to overcome the bottleneck of 3D annotated data. Early zero-shot methods \cite{li2025seeground,jin2025spazer} still relied on the input paradigm of pre-reconstructed 3D point clouds, simplifying the task to candidate region matching.

In recent years, VLM-Grounder \cite{xu2025vlm} and TAB \cite{wang2026think} have pioneered the exploration of technical approaches for 3D localization directly from 2D image sequences. Among these, VLM-Grounder associates objects across different views through multi-view semantic matching; however, due to the lack of guidance from deterministic 3D geometric constraints, its performance drops sharply under extreme view changes. TAB is the first to reformulate 3D-VG as a process of active semantic reasoning and geometric reconstruction; however, it remains essentially a unidirectional, serial pipeline, suffering from inherent flaws such as unidirectional error accumulation and insufficient robustness. In contrast, our DACA framework proposed in this paper fundamentally resolves these issues through a dual-anchor collaboration mechanism.

\noindent\textbf{VLM-based tool-calling agents.} The rapid development of LLMs and VLMs has given rise to the paradigm of tool-calling agents. ReAct \cite{yao2023react} first introduced the Reasoning\&Acting loop, enabling agents to interact with the environment by invoking external tools to perform complex reasoning tasks. In the visual domain, Voyager \cite{wang2023voyager} and GPT-4V \cite{hurst2024gpt} demonstrated the powerful capabilities of VLM agents in embodied navigation and visual reasoning. However, most VLM agents generally adopt a single serial architecture, relying on general-purpose modules to complete tasks, and lack decoupled and collaborative tools for specialized functions.

In the 3D-VG task, semantic reasoning and geometric perception impose different demands on modeling capabilities, making it difficult for a single general-purpose module to effectively handle both types of tasks simultaneously. To address this, this paper introduces a dual-anchor collaboration mechanism within a single agent. By leveraging the division of labor and collaborative optimization between two specialized tools—semantic anchors and geometric anchors—we achieve superior semantic understanding and geometric localization performance compared to traditional agents.

\noindent\textbf{VLMs for 3d understanding.} Current 3D VLMs can be broadly categorized into two types: one uses specialized 3D encoders to directly process point clouds such as PointLLM \cite{xu2024pointllm} and SpatialLM \cite{mao2025spatiallm}; the other aggregates multi-view 2D image features into unified 3D spatial features such as 3D-LLM \cite{hong20233d} and LLaVA-3D \cite{zhu2025llava}. Although these models demonstrate powerful scene-level reasoning capabilities, they require resource-intensive cross-modal alignment training and rely on pre-reconstructed 3D inputs.

Different from the aforementioned methods, the proposed DACA does not require 3D-specific pre-training. We leverage the reasoning capabilities of 2D VLMs within the agent loop and achieve precise 3D spatial understanding directly from raw RGB-D streams by combining the fundamental model with multi-view geometry and TSDF (Truncated Signed Distance Function) voxelization.

\section{Method}
\subsection{Overview}\label{label:3.1}
Given a natural language query \(Q\), a continuous RGB-D image sequence \(S = \{(I_i, D_i)\}_{i=1}^T \) consisting of \(T\) frames (where \(I_i\) is the RGB image of the \( i \)-th frame and \(D_i\) is the corresponding depth image), as well as the camera intrinsic matrix \(K_\text{cam}\) and the extrinsic matrix \(T_\text{c2w}\), the proposed DACA aims to accurately output the axis-aligned 3D bounding box \(B \in \mathbb{R}^6 \) of the target under zero-shot settings. Specifically, DACA reformulates the 3D-VG task as an adaptive closed-loop system that integrates semantic reasoning and geometric perception, driven by both semantic and geometric anchors. As shown in Figure \ref{fig2}, the overall agent pipeline proceeds as follows:

\noindent\textbf{1. Dual anchor initialization}: First, the semantic anchor is initialized by extracting the target’s class, visual attributes, and spatial constraint through a query resolution module. Simultaneously, the target’s 2D instance mask is extracted from the selected optimal reference frame, and the initial 3D centroid is obtained via camera parameter backprojection to initialize the geometric anchor.

\noindent\textbf{2. Two-stage progressive extension}: We then utilize RAFT optical flow \cite{teed2020raft} to perform local bidirectional tracking near the reference frame, replacing VLM’s inefficient frame-by-frame global semantic validation and improving the efficiency of view acquisition while maintaining tracking accuracy. Moreover, based on TSDF-voxelized \cite{werner2014truncated} global scene visibility, we actively search for distant views that differ from the reference viewpoint, which addresses the limitation of traditional geometric extension that only covers adjacent viewpoints.

\noindent\textbf{3. Dual-anchor cross-validation}: A dual-anchor cross-validation mechanism is introduced to filter unreliable views. Semantic anchors verify the semantic consistency of objects in views obtained through geometric extension to filter out false positives, while geometric anchors verify the spatial consistency of views obtained through semantic tracking to eliminate semantic confusion.

\noindent\textbf{4. 3D reconstruction and bounding box estimation}: Finally, we fuse the 2D instance masks from all validated images, reconstruct the complete 3D point cloud of the target through 2D-3D matrix transformation and voxel fusion, and compute the precise axis-aligned 3D bounding box.

\subsection{Semantic Anchor Tool}\label{sec3.2}
Given that the semantic temporal extension (STE) in previous methods \cite{wang2026think,liu2025mesh} adopts a paradigm of frame-by-frame full-image VLM verification, it not only suffers from low computational efficiency but also tends to produce false positives in full-image semantic matching when the object is occluded or the viewpoint changes, thereby limiting the coverage of temporal extension and the overall operational efficiency of the agent. 

To address these shortcomings, we propose a motion-guided semantic temporal extension mechanism (MG-STE), which deeply integrates the motion prior provided by dense optical flow with the local semantic verification of the VLM. The core design philosophy of this mechanism lies in fully leveraging the inherent spatiotemporal continuity of video sequences. By using optical flow to predict the approximate spatial location of the target in adjacent frames, we compress the VLM’s attention from the entire image to the local region where the target is located, thereby significantly reducing computational overhead while ensuring the accuracy of semantic verification.

Specifically, given a reference frame \(I_\text{ref}\), a reference mask \(M_\text{ref} \in \{0,1\}^{H \times W}\), and its time index \(t_\text{ref}\), the semantic anchor agent initiates a bidirectional tracking loop along the timeline. For each adjacent candidate frame \(I_t\), the dense optical flow field \(\mathbf{F}_{\text{ref} \to t} \in \mathbb{R}^{H \times W \times 2}\) between the two frames is first computed using the RAFT optical flow estimator, where the displacement vector for each pixel \((x,y)\) is \((F_x(x,y), F_y(x,y))\). Further, the reference mask is projected onto the current frame to obtain the predicted mask of the object. This process can be expressed as:
\begin{equation}
	\begin{aligned}
		\mathbf{G}(x,y) &= [x + F_x(x,y), y + F_y(x,y)], \\
		&\quad \forall (x,y) \in [0,W-1] \times [0,H-1]
	\end{aligned}
\end{equation}
where \(\mathbf{G}(x,y)\) denotes the corresponding coordinates in the target frame of the reference frame pixel $(x,y)$. The projected mask is then obtained using bilinear interpolation and binarized:
\begin{equation}
	M_{\text{warp}}(x,y) = \left\{\begin{array}{@{}l@{\,}l@{}} 
		1, & \text{if} \ \texttt{grid\_sample}(M_\text{ref}, \mathbf{G})(x,y) > 0.5 \\
		0, & \text{otherwise}
	\end{array}\right.
\end{equation}
In particular, $\texttt{grid\_sample}$ uses bilinear interpolation, zero-padding for boundaries, and corner-alignment parameters to ensure flow accuracy. Subsequently, the VLM is invoked only within the bounding box of the prediction mask \(M_{\text{warp}}\) to verify semantic consistency.

If the verification passes, SAM3 \cite{carion2025sam} is invoked to generate a precise instance mask for the current frame, and this frame, along with its corresponding depth image and mask, is added to the semantic tracking segment. Starting from the reference frame, we expand $2 \times K_{\text{half}}$ steps in both directions along the timeline—backward (\(t < t_{\text{ref}}\) ) and forward (\(t > t_{\text{ref}}\) )—to construct a complete semantic tracking segment \(\mathcal{S}_{\text{sem}}\). The backward and forward tracking set is defined as:
\begin{equation}
	\begin{aligned}
		\mathcal{S}_{\text{backward}} &= \cup_{k=1}^{K_{\text{half}}} \{ (I_{t_{\text{ref}}-k}, D_{t_{\text{ref}}-k}, M_{t_{\text{ref}}-k}) \}, \\
		\mathcal{S}_{\text{forward}} &= \cup_{k=1}^{K_{\text{half}}} \left\{ (I_{t_{\text{ref}}+k}, D_{t_{\text{ref}}+k}, M_{t_{\text{ref}}+k}) \right\}
	\end{aligned}
\end{equation}

If the above verification fails, a collaboration mechanism with the sementic and geometric anchor tools is triggered to request continued tracking of the target via geometric projection, thereby avoiding interruption:
\begin{equation}
	(I_t, M_t)_{sem} = \mathbb{I}\left(\texttt{VLM\_Score}(I_t, M_t, Q) \geq \tau_{\text{sem}} \right)
\end{equation}
\begin{equation}
	(I_t, M_t)_{{geo}} = \mathbb{I}\left( \| \text{Centroid}(M_t) - \pi(P_{\text{geo}}, T_{c2w}^t) \|_2 \leq \tau_{\text{geo}} \right)
\end{equation}
where $\texttt{VLM\_Score}$ represents a semantic matching score function, \(\mathbb{I}(\cdot)\) is the indicator function, \(\tau_{\text{sem}}\) is the semantic validation threshold, \(P_{\text{geo}}\) is the 3D centroid of the geometric anchor, \(\pi(\cdot)\) is the projection function from a 3D point to a 2D pixel, and \(\tau_{\text{geo}}\) is the spatial distance threshold.

\subsection{Geometric Anchor Tool}\label{sec3.3}
Traditional multi-view geometric extension (MGE) relies on a single 3D center of mass for projection \cite{wang2026think}, which fails to capture the spatial extent and shape uncertainty of the object. To address this, we propose keypoint‑aware multi‑view geometric extension (KA‑MGE), which upgrades the single‑point geometric constraint to a three‑dimensional geometric distribution of nine keypoints. Combined with TSDF voxel ray tracing to achieve precise visibility determination, this approach robustly retrieves valid views of observable targets across the entire image sequence.

Given the initial \(\mathcal{S}_{\text{sem}}\) output by the semantic anchor, the geometric anchor first back‑projects the pixels within the masked regions of each frame into the 3D world coordinate system, constructs the scene TSDF voxel grid \(\mathcal{S}_{\text{tsdf}}\), and extracts the target point cloud \({P} \in \mathbb{R}^{N \times 3}\) from it. Based on this point cloud, an alignment bounding box \({B} \in \mathbb{R}^6\) is constructed, and 9 keypoints \({K} = \{\mathbf{k}_i\}_{i=1}^9\) are generated, comprising 8 box vertices and 1 centroid, collectively forming the 3D geometric distribution of the target to explicitly model the uncertainty in the target’s spatial position.

For any candidate frame \(I_t\) in the image sequence, we first transform the pixel points \((u,v)\) from the depth‑aligned camera coordinates \(p_c\) to the original world coordinate:
\begin{equation}
	\boldsymbol{p}_c = D_{u,v} \cdot K_{\text{cam}}^{-1} \begin{bmatrix} u \\ v \\ 1 \end{bmatrix}
	= D_{u,v} \cdot \begin{bmatrix} \dfrac{u - c_x}{f_x} \\[4pt] \dfrac{v - c_y}{f_y} \\[4pt] 1 \end{bmatrix},
\end{equation}

\begin{equation}
	K_{\text{cam}} = \begin{bmatrix} f_x & 0 & c_x \\ 0 & f_y & c_y \\ 0 & 0 & 1 \end{bmatrix}
\end{equation}
where \(D_{u,v}\) indicates the actual physical depth, \(K_{\text{cam}}\) denotes the intrinsic matrix of the camera, \(f_x\) and \(f_y\) represent the focal lengths, and \(c_x\) and \(c_y\) represent the coordinates of the principal point, respectively.

Subsequently, through the camera‑to‑camera transformation and the transformation to the world coordinate system, we finally obtain the aligned 3D world coordinate point \(p_{w}\):
\begin{equation}
	\begin{bmatrix} \boldsymbol{p}_w \\ 1 \end{bmatrix}
	= T_{\text{c2w}} \begin{bmatrix} \boldsymbol{p}_c \\ 1 \end{bmatrix}
	= \begin{bmatrix} \mathbf{R} & \mathbf{t} \\ \mathbf{0}^\mathrm{T} & 1 \end{bmatrix}
	\begin{bmatrix} \boldsymbol{p}_c \\ 1 \end{bmatrix}.
\end{equation}

Next, we perform ray‑tracing visibility checks based on the TSDF‑based voxel \(\mathcal{V}_{\text{tsdf}}\):
\begin{equation}
	v_i = \texttt{is\_visible}(\mathcal{V}_{\text{tsdf}}, K_{\text{cam}}, T_{\text{c2w}})
\end{equation}
where \(\texttt{is\_visible}\) is used to determine whether a 3D point is occluded from the camera's viewpoint, and \(v_i \in \{0,1\}\) indicates whether the \(i\)-th keypoint is visible.

If the number of valid visible keypoints satisfies \(\sum_{i=1}^9 v_i \ge 3\), then the current frame is determined to be observable and included in the candidate set. The mean of all visible projection points is used as the point prompt for SAM3 to perform precise instance segmentation within the target region:
\begin{equation}\label{eq9}
	{c} = \frac{1}{\sum_{i=1}^9 v_i} \sum_{i=1}^9 v_i \cdot {p}_i
\end{equation}

During the mask selection phase, we prioritize masks that contain the projection point \({c}\). If no such mask exists, we select the mask closest to the projection point \({c}\):
\begin{equation}
	M_t^* = \mathop{\arg\min}_{M_t} \, \min_{(x,y)\in M_t} \bigl\| (x,y) - c \bigr\|_2
\end{equation}

As a result, the validated frame and its mask are added to the global geometric view set \(\mathcal{S}_{\text{geo}}\).

We argue that replacing depth value approximation with TSDF‑based ray‑tracing visibility checks reduces the occlusion misclassification. Additionally, KA‑MGE traverses all video frames globally, effectively addressing the long‑range and wide‑angle gaps that semantic tracking cannot cover.
\begin{table*}[t]
	\centering
	\caption{Comparison with state-of-the-art methods on ScanRefer. Note that ``w/ PC'' denotes methods that require 3D point clouds as input.}
	\vspace{-10pt}
	\label{tab1}
	\resizebox{\linewidth}{!}{
		\begin{tabular}{l|c|c|cc|cc|cc}
			\toprule
			\multirow{2}{*}{Method} & \multirow{2}{*}{LLM/VLM} & \multirow{2}{*}{w/ PC} & \multicolumn{2}{c}{Unique} & \multicolumn{2}{c}{Multiple} & \multicolumn{2}{c}{Overall} \\
			\cmidrule(lr){4-5} \cmidrule(lr){6-7} \cmidrule(lr){8-9}
			& & & Acc@0.25 & Acc@0.5 & Acc@0.25 & Acc@0.5 & Acc@0.25 & Acc@0.5 \\
			\midrule
			\rowcolor{blue!10} \multicolumn{9}{c}{Fully Supervised Methods (One/Two-Stage)} \\
			\midrule
			ScanRefer \cite{chen2020scanrefer} & -- & \deepgreen{\Checkmark} & 67.6 & 46.2 & 32.1 & 21.3 & 39.0 & 26.1 \\
			3DVG-Transformer \cite{zhao20213dvg} & -- & \deepgreen{\Checkmark} & 81.9 & 60.6 & 39.3 & 28.4 & 47.6 & 34.7 \\
			BUTD-DETR \cite{jain2022bottom} & -- & \deepgreen{\Checkmark} & 84.2 & 66.3 & 46.6 & 35.1 & 52.2 & 39.8 \\
			EDA \cite{wu2023eda} & -- & \deepgreen{\Checkmark} & 85.8 & 68.6 & 49.1 & 37.6 & 54.6 & 42.3 \\
			G3-LQ \cite{wang2024g} & -- & \deepgreen{\Checkmark} & 88.6 & 73.3 & 50.2 & 39.7 & 56.0 & 44.7 \\
			MCLN \cite{qian2024multi} & -- & \deepgreen{\Checkmark} & 86.9 & 72.7 & 52.0 & 40.8 & 57.2 & 45.5 \\
			\midrule
			\rowcolor{orange!10}\multicolumn{9}{c}{Fully Supervised Methods (LLM/VLM-Based)} \\
			\midrule
			SPAR-mix \cite{zhang2026flatland} & InternVL-2.5-8B & \deepred{\XSolidBrush}/\deepgreen{\Checkmark} & -- & -- & -- & -- & \deepred{31.9}/\deepgreen{48.8} & \deepred{12.4}/\deepgreen{43.1} \\
			LLaVA-3D \cite{zhu2025llava} & LLaVA-Video-7B & \deepred{\XSolidBrush} & -- & -- & -- & -- & 50.1 & 42.7 \\
			Chat-Scene \cite{huang2024chat} & Vicuna-7B & \deepgreen{\Checkmark} & 89.5 & 82.4 & 47.7 & 42.9 & 55.5 & 50.2 \\
			VG-LLM \cite{zheng2026learning} & Qwen2.5-VL-7B & \deepred{\XSolidBrush}/\deepgreen{\Checkmark} & -- & -- & -- & -- & \deepred{41.6}/\deepgreen{57.6} & \deepred{14.9}/\deepgreen{50.9} \\
			Video-3D-LLM \cite{zheng2025video} & LLaVA-Video-7B & \deepgreen{\Checkmark} & 87.9 & 78.3 & 50.9 & 45.3 & 58.1 & 51.7 \\
			GPT4Scene \cite{qi2025gpt4scene} & Qwen2-VL-7B & \deepgreen{\Checkmark} & 90.3 & 83.7 & 56.4 & 50.9 & 62.6 & 57.0 \\
			3D-RS \cite{huang20263drs} & LLaVA-Next-Video & \deepgreen{\Checkmark} & 87.4 & 77.9 & 57.0 & 50.8 & 62.9 & 56.1 \\
			\midrule
			\rowcolor{purple!10}\multicolumn{9}{c}{{Zero-Shot Methods}} \\
			\midrule
			LLM-Grounder \cite{yang2024llm} & GPT-4 turbo & \deepgreen{\Checkmark} & -- & -- & -- & -- & 17.1 & 5.3 \\
			ZS3DVG \cite{yuan2024visual} & GPT-4 turbo & \deepgreen{\Checkmark} & 63.8 & 58.4 & 27.7 & 24.6 & 36.4 & 32.7 \\
			SeeGround \cite{li2025seeground} & Qwen2-VL-72B & \deepgreen{\Checkmark} & 75.7 & 68.9 & 34.0 & 30.0 & 44.1 & 39.4 \\
			UniGround \cite{zhang2026uniground} & GPT-5 & \deepred{\XSolidBrush} & -- & -- & -- & -- & 46.1 & 34.1 \\
			VLM-Grounder \cite{xu2025vlm} & GPT-4o & \deepred{\XSolidBrush} & 66.0 & 29.8 & 48.3 & 33.5 & 51.6 & 32.8 \\
			SeqVLM \cite{lin2025seqvlm} & Doubao-1.5-pro & \deepgreen{\Checkmark} & 77.3 & 72.7 & 47.8 & 41.3 & 55.6 & 49.6 \\
			SPAZER \cite{jin2025spazer} & GPT-4o & \deepgreen{\Checkmark} & 80.9 & 72.3 & 51.7 & 43.4 & 57.2 & 48.8 \\
			TAB \cite{wang2026think} & Qwen3-VL-32B & \deepred{\XSolidBrush}/\deepgreen{\Checkmark} & \deepred{90.2}/\deepgreen{90.2}& \deepred{57.6}/\deepgreen{77.2} & \deepred{60.1}/\deepgreen{60.8} & \deepred{39.9}/\deepgreen{52.5} & \deepred{71.2}/\deepgreen{71.6} & \deepred{46.4}/\deepgreen{61.6} \\
			\midrule
			\rowcolor{gray!10} \textbf{DACA (Ours)} & Qwen3-VL-32B & \deepred{\XSolidBrush}/\deepgreen{\Checkmark} & \deepred{\textbf{91.3}}/\deepgreen{92.4} & \deepred{\textbf{59.8}}/\deepgreen{79.3} & \deepred{\textbf{63.3}}/\deepgreen{63.9} & \deepred{\textbf{46.2}}/\deepgreen{54.4} & \deepred{\textbf{73.6}}/\deepgreen{74.4} & \deepred{\textbf{51.2}}/\deepgreen{63.6} \\
			\bottomrule
		\end{tabular}
	}
\end{table*}

\subsection{3D Reconstruction and Box Estimation}\label{sec3.4}
Based on the final merged global mask set \(\mathcal{S}_{\text{geo}}\), DACA first combines all pixels within the valid mask regions with their corresponding depth maps and camera parameters, then back-projects them into the 3D world coordinate system to construct the initial point cloud of the object. Following TAB, we sequentially apply a statistical outlier removal algorithm \cite{rusu2010semantic} and the DBSCAN density-based clustering algorithm \cite{ester1996density} to perform multi-stage filtering on the initial point cloud, removing outliers and extracting the target’s main geometric structure. Finally, by computing the spatial extrema of the filtered, clean point cloud, we obtain the target’s axis-aligned 3D bounding box \(B \in \mathbb{R}^6\), thereby completing the entire zero-shot 3D visual grounding process.

\section{Experiments}
\subsection{Experimental Setup}\label{label:4.1}
\textbf{Datasets.} We conduct experiments on two benchmarks for 3D visual grounding: ScanRefer \cite{chen2020scanrefer} and Nr3D \cite{achlioptas2020referit3d}, both derived from ScanNet \cite{dai2017scannet} scenes.

ScanRefer has annotated 51,583 manually labeled query-target object pairs for ScanNet. These queries are categorized into \textit{Unique} (only one instance of the target class in the scene) and \textit{Multiple} (multiple same-class distractors requiring fine-grained discrimination). 

Part of the ReferIt3D \cite{achlioptas2020referit3d} dataset, the Nr3D contains 41,503 queries for ScanNet scenes, collected via a two-player reference game. Each target has at least one distractor, with the \textit{Easy} subset containing one distractor and \textit{Hard} containing two or more. Queries are further divided into \textit{View-Dependent} and \textit{View-Independent} subsets.

The main evaluations use 250 validation queries per dataset, following the subset protocol of VLM-Grounder and SPAZER \cite{xu2025vlm,jin2025spazer}. The dataset loader is configured for up to 300 ScanNet RGB-D frames per sequence. The agent separately enumerates scene frames, so the frame list supplied to the paired rerun must be frozen and logged. The evaluation script reads TAB's refined manifests \cite{wang2026think}. Their released TAB predictions match by scene, target, and query, with multiplicities. The manifests contain 92 Unique and 158 Multiple queries across 115 ScanRefer scenes, and 250 query rows across 99 Nr3D scenes.

\noindent\textbf{Evaluation metrics.} For ScanRefer, we report Acc@0.25 and Acc@0.5 as the primary metrics across Unique, Multiple, and Overall splits. For Nr3D, we report top-1 accuracy across Easy, Hard, View-Dependent (Dep.), View-Independent (Indep.), and Overall splits. Notably, DACA can optionally use pre-processed 3D point clouds as input (denoted as ``w/ PC'').

\noindent\textbf{Implementation details.} DACA is built entirely upon open-source models. We employ Qwen3-VL-32B \cite{bai2025qwen3} as the primary VLM agent for semantic reasoning and tool orchestration. For foundational vision tools, we utilize RAFT \cite{teed2020raft} for dense optical flow estimation and SAM3 \cite{carion2025sam} for object instance segmentation. The semantic temporal extension and geometric extension processes are each limited to a maximum of 16 frames in both forward and backward directions. The TSDF voxel size is set to 0.02m. The semantic verification threshold $\tau_{\text{sem}}$ and spatial consistency threshold $\tau_{\text{geo}}$ are set to 0.5 and 0.4m, respectively. All experiments are conducted on 8$\times$NVIDIA RTX 5090 GPUs. 

\subsection{Comparison with State-of-the-Art Methods}
\textbf{Results on ScanRefer.} Table \ref{tab1} presents the comparison results on the ScanRefer benchmark. DACA achieves an overall Acc@0.25 of 73.6\% and Acc@0.5 of 51.2\%, establishing a new state-of-the-art among zero-shot methods. Notably, DACA significantly outperforms all prior zero-shot approaches, including those that rely on pre-processed 3D point clouds as strong structural priors. Compared to the most competitive zero-shot baseline TAB \cite{wang2026think}, which also operates directly on RGB-D streams, DACA achieves a substantial improvement of +2.4\%/+4.8\% in overall Acc@0.25/Acc@0.5. This gain is primarily attributed to the dual-anchor collaborative mechanism, which effectively addresses the unidirectional error accumulation inherent in TAB's pipeline.

\noindent\textbf{Results on Nr3D.} Table \ref{tab2} summarizes the comparison on the Nr3D benchmark. DACA achieves an overall accuracy of 70.4\%, substantially outperforming all prior zero-shot methods. Compared to TAB (68.0\%), which also forgoes 3D point cloud inputs, DACA attains a +2.4\% improvement. The gain is consistent across all evaluation splits, with particularly notable advances on the challenging \textit{Hard} (+3.5\%) and \textit{View-Independent} (+2.6\%) subsets. These results validate that DACA's keypoint-aware multi-view geometric extension and dual-anchor mechanism effectively handle complex spatial relationships and viewpoint-dependent descriptions.

It is important to note that prior zero-shot methods such as SeqVLM and SPAZER benefit from ground-truth 3D bounding box proposals provided by the ScanRefer and Nr3D benchmarks, which serve as strong priors that substantially simplify the grounding task. In contrast, DACA operates strictly without such priors, directly reconstructing the target's 3D geometry from raw RGB-D observations. Despite this more challenging setting, DACA surpasses all zero-shot baselines and even outperforms several fully supervised methods, which was trained on extensive annotated 3D data.

\begin{table}[t]
	\centering
	\caption{Comparison with state-of-the-art methods on Nr3D. 
	}
	\label{tab2}
	\vspace{-10pt}
	\resizebox{\linewidth}{!}{%
		\begin{tabular}{l|c|c|cc|cc|c}
			\toprule
			Method & LLM/VLM & w/ PC & Easy & Hard & Dep. & Indep. & Overall \\
			\midrule
			\rowcolor{blue!10} \multicolumn{8}{c}{Fully Supervised Methods} \\
			\midrule
			ReferIt3DNet \cite{achlioptas2020referit3d} & -- & \deepgreen{\Checkmark} & 43.6 & 27.9 & 32.5 & 37.1 & 35.6 \\
			InstanceRefer \cite{yuan2021instancerefer} & -- & \deepgreen{\Checkmark} & 46.0 & 31.8 & 34.5 & 41.9 & 38.8 \\
			3DVG-Transformer \cite{zhao20213dvg} & -- & \deepgreen{\Checkmark} & 48.5 & 34.8 & 34.8 & 43.7 & 40.8 \\
			EDA \cite{wu2023eda} & -- & \deepgreen{\Checkmark} & 58.2 & 46.1 & 50.2 & 53.1 & 52.1 \\
			BUTD-DETR \cite{jain2022bottom} & -- & \deepgreen{\Checkmark} & 54.6 & 60.7 & 48.4 & 46.0 & 58.0 \\
			SceneVerse \cite{jia2024sceneverse} & -- & \deepgreen{\Checkmark} & 72.5 & 57.8 & 56.9 & 67.9 & 64.9 \\
			\midrule
			\rowcolor{purple!10}\multicolumn{8}{c}{{Zero-Shot Methods}} \\
			\midrule
			ZS3DVG \cite{yuan2024visual} & GPT-4 turbo & \deepgreen{\Checkmark} & 46.5 & 31.7 & 36.8 & 40.0 & 39.0 \\
			SeeGround \cite{li2025seeground} & Qwen2-VL-72B & \deepgreen{\Checkmark} & 54.5 & 38.3 & 42.3 & 48.2 & 46.1 \\
			VLM-Grounder \cite{xu2025vlm} & GPT-4o & \deepred{\XSolidBrush} & 55.2 & 39.5 & 45.8 & 49.4 & 48.0 \\
			SeqVLM \cite{lin2025seqvlm} & Doubao-1.5-pro & \deepgreen{\Checkmark} & 58.1 & 47.4 & 51.0 & 54.5 & 53.2 \\
			SPAZER \cite{jin2025spazer} & GPT-4o & \deepgreen{\Checkmark} & 65.0 & 62.5 & 61.0 & 66.0 & 63.8 \\
			TAB \cite{wang2026think} & Qwen3-VL-32B & \deepred{\XSolidBrush} & {72.1} & {63.2} & {62.5} & {71.4} & {68.0} \\
			\midrule
			\rowcolor{gray!10} \textbf{DACA (Ours)} & Qwen3-VL-32B & \deepred{\XSolidBrush} & \textbf{73.5} & \textbf{66.7} & \textbf{64.6} & \textbf{74.0} & \textbf{70.4} \\
			\bottomrule
		\end{tabular}%
	}
\end{table}
\begin{figure}
	\centering
	\includegraphics[width=\linewidth]{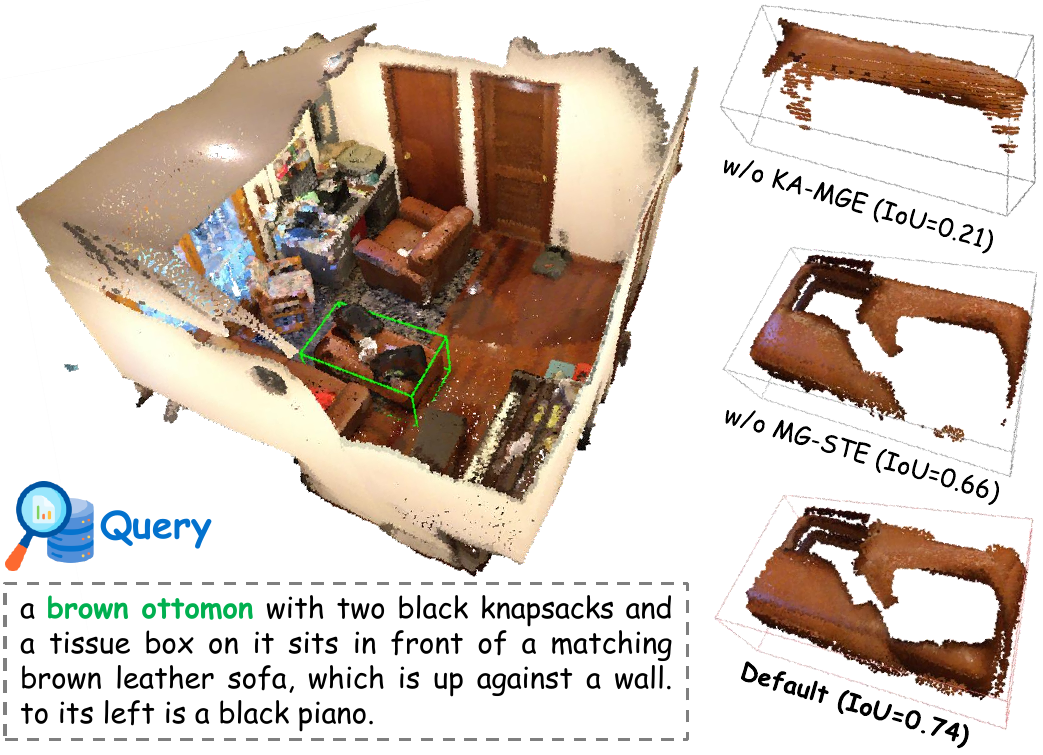}
	\caption{Qualitative comparison of the MG-STE and KA-MGE components within DACA.}
	\label{fig3}
\end{figure}
\subsection{Ablation Studies}
We conduct comprehensive ablation studies to validate the effect of each component for DACA, as shown in Table \ref{tab3}. All experiments are performed on the ScanRefer validation subset, with overall results reported.
\begin{table}
	\centering
	\caption{Ablation studies on ScanRefer.}
	\vspace{-10pt}
	\resizebox{0.9\linewidth}{!}{
		\begin{tabular}{lcc}
			\toprule
			Configuration & Acc@0.25 & Acc@0.5 \\
			\midrule
			\multicolumn{3}{c}{\textbf{(a) Component Analysis}} \\
			\midrule
			Single-frame baseline & 42.4 & 23.2 \\
			+ MG-STE & 58.4 & 35.2 \\
			+ KA-MGE & 60.0 & 38.4 \\
			\rowcolor{gray!10} + MG-STE + KA-MGE  & \textbf{73.6} & \textbf{51.2} \\
			\midrule
			\multicolumn{3}{c}{\textbf{(b) Cross-Validation Analysis}} \\
			\midrule
			w/o semantic$\rightarrow$geometric validation & 72.0 & 50.4 \\
			w/o geometric$\rightarrow$semantic validation & 71.6 & 50.0 \\
			w/o both cross-validation & 70.4 & 47.6 \\
			\rowcolor{gray!10} Full DACA & \textbf{73.6} & \textbf{51.2} \\
			\midrule
			\multicolumn{3}{c}{\textbf{(c) Number of Keypoints in KA-MGE}} \\
			\midrule
			1 keypoint (centroid only) & 71.6 & 47.6 \\
			\rowcolor{gray!10} 9 keypoints (8 corners + 1 centroid) & \textbf{73.6} & \textbf{51.2} \\
			13 keypoints (+4 face centers) & 72.8 & 48.8 \\
			\midrule
			\multicolumn{3}{c}{\textbf{(d) VLM Backbone}} \\
			\midrule
			Qwen3-VL-4B & 58.4 & 47.2 \\
			Qwen3-VL-8B & 62.4 & 49.6 \\
			\rowcolor{gray!10} Qwen3-VL-32B & \textbf{73.6} & \textbf{51.2} \\
			\bottomrule
		\end{tabular}
	}
	\label{tab3}
\end{table}

\noindent\textbf{Effect of components.} As shown in Table \ref{tab3}(a), we integrate each proposed module based on the single-frame baseline. The baseline only obtains 42.4\% Acc@0.25 and 23.2\% Acc@0.5, which proves that single-frame reconstruction lacks temporal and geometric cues and suffers severe performance bottlenecks. Equipping the baseline with MG-STE alone lifts Acc@0.25 to 58.4\%, showing that motion-guided temporal propagation can exploit cross-frame temporal information. Moreover, adding KA-MGE separately brings Acc@0.25 up to 60.0\%, which verifies the keypoint-based geometric extension is beneficial. Combining KA-MGE and MG-STE together reaches the peak value 73.6\% Acc@0.25 and 51.2\% Acc@0.5, which indicates the two modules complement each other: temporal semantic propagation and geometric spatial constraints jointly eliminate reconstruction ambiguity and greatly boost location accuracy.

\noindent\textbf{Effect of dual-anchor cross-validation.} Table \ref{tab3}(b) analyzes the effectiveness of bidirectional cross-validation in the full DACA pipeline. Removing semantic$\rightarrow$geometric cross-validation slightly degrades the metric to 72.0\%, because unfiltered geometric matching false samples are introduced into the fusion process. Disabling geometric$\rightarrow$semantic validation leads to a mild drop to 71.6\%, caused by uncorrected semantic feature mismatches. If both cross-validation branches are removed simultaneously, Acc@0.25 falls sharply to 70.4\%. The obvious performance decline after removing either or both validation branches demonstrates that the dual-anchor bidirectional consistency check is an indispensable part of our framework to suppress matching noise.

\noindent\textbf{Effect of the number of keypoints in KA-MGE.} Table \ref{tab3}(c) ablates the keypoint configuration inside KA-MGE. Only using object centroid as the single keypoint gets 71.6\% Acc@0.25. Expanding keypoints to eight corners plus one centroid (9 keypoints) improves the metric to the optimal 73.6\%. Further increasing keypoints to 4 extra face centers (13 keypoints) slightly reduces accuracy to 72.8\%. The results show that 9 keypoints can fully cover the spatial boundary of objects and provide sufficient geometric constraints, while redundant keypoints introduce redundant calculation and slight matching noise instead of performance gain.

\noindent\textbf{Effect of VLM backbone.} Table \ref{tab3}(d) evaluates DACA's performance equipped with VLM backbones of different parameter scales. The lightweight Qwen3-VL-4B only attains 58.4\% Acc@0.25. Scaling up to Qwen3-VL-8B lifts Acc@0.25 to 62.4\%, benefiting from strengthened vision-language comprehension. Our default Qwen3-VL-32B achieves the optimal 73.6\% Acc@0.25 with robust joint semantic-geometric reasoning. Although our DACA can partially remedy insufficient spatial perception of tiny VLMs, extremely lightweight backbones (such as Qwen3-VL-2B) still suffer fundamental degradation due to poor action parsing and tool calling capacity.
\begin{figure}[t]
	\centering
	\includegraphics[width=\linewidth]{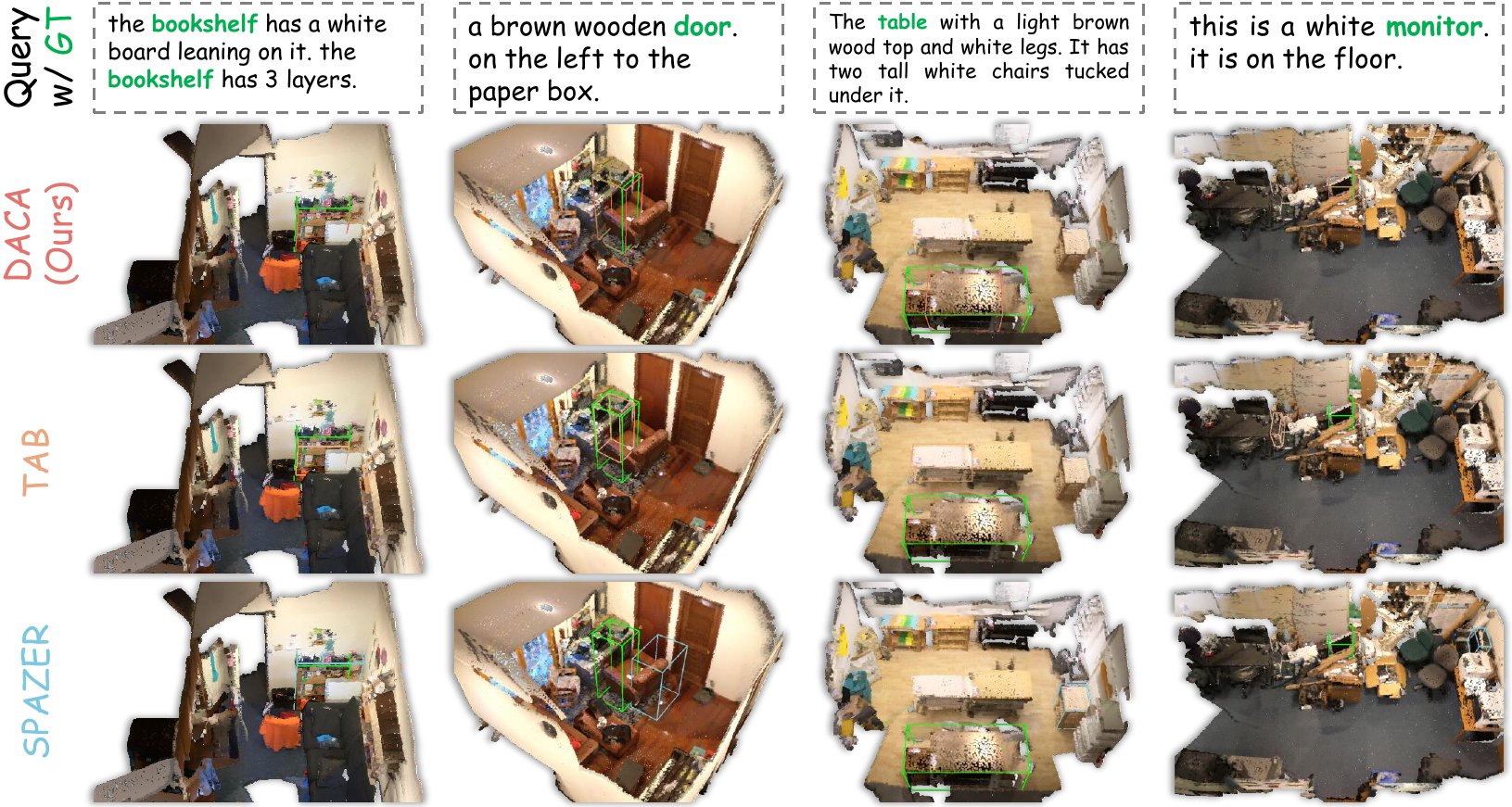}
	\caption{Qualitative comparison of zero-shot 3D-VG prediction results on ScanRefer. GT bounding boxes are marked in \textcolor[HTML]{00B050}{green}. The left shows the \textit{unique} type, and the right shows the \textit{multiple} type.}
	\vspace{-10pt}
	\label{fig4}
\end{figure}
\begin{figure}[t]
	\centering
	\includegraphics[width=\linewidth]{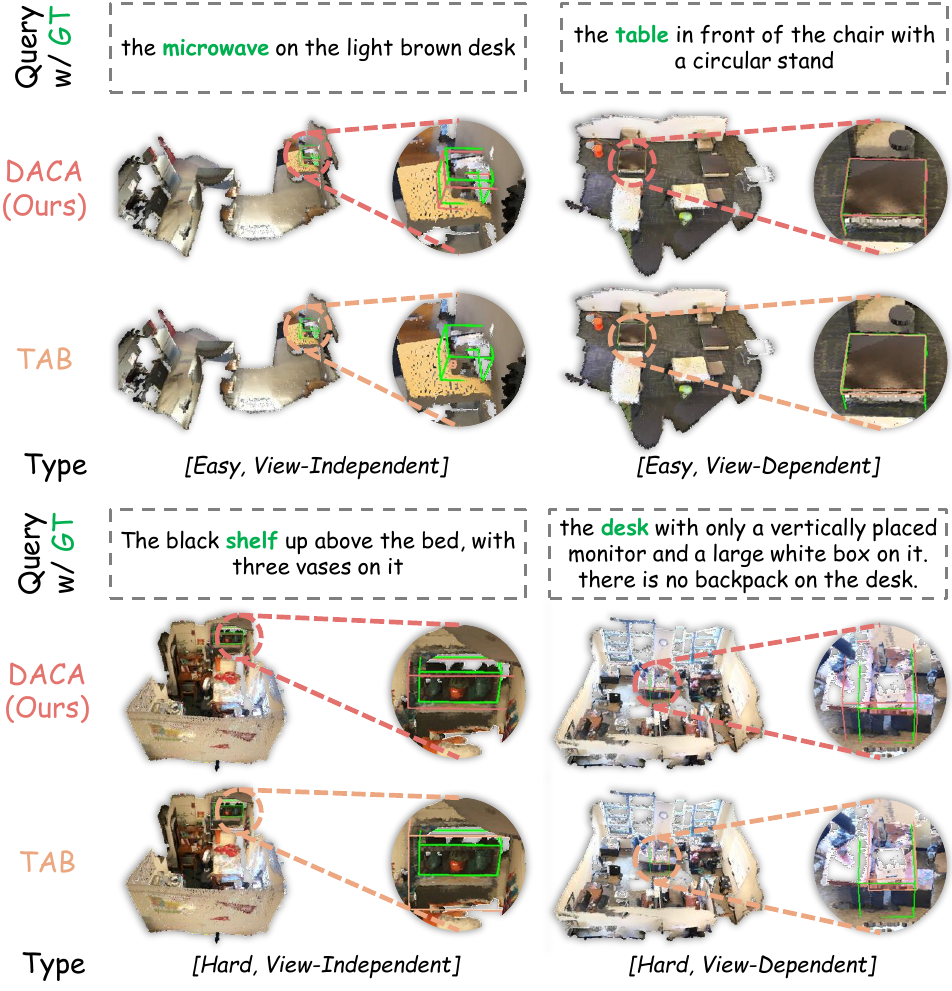}
	\caption{Qualitative comparison of zero-shot 3D-VG prediction results on Nr3D. GT bounding boxes are marked in \textcolor[HTML]{00B050}{green}.}
	\vspace{-10pt}
	\label{fig5}
\end{figure}

\subsection{Qualitative Analysis}
Figures \ref{fig4} and \ref{fig5} present qualitative comparisons between DACA and the competitive zero-shot baselines across a number of representative challenging scenarios. Further observation reveals the following points.

When faced with \textbf{occlusion scenarios}, DACA leverages keypoint projection and TSDF tracing to fuse multi-view observations and reconstruct the complete object, effectively avoiding the bounding box deviations caused by incomplete masks and projection errors. When similar \textbf{interfering objects} are present in a scene, dual-anchor cross-validation verifies spatial consistency, filters out false positives, and accurately pinpoints the target. For query descriptions with \textbf{broad constraints}, global visibility overcomes the limitations of local fields of view, offering greater adaptability. In challenging \textbf{multi-target detection} tasks, the agent’s robustness against interference and recognition stability are enhanced through geometric-semantic verification.

\section{Conclusion}
This paper proposes DACA, a zero-shot 3D-VG agent that establishes a new paradigm for collaborative reasoning between semantic and geometric anchors. DACA achieves precise multi-view object tracking by leveraging semantic anchor motion to guide semantic temporal expansion, and ensures robust spatial perception via keypoint geometric expansion and TSDF visibility derived from geometric anchors. Through bidirectional closed-loop collaborative optimization of the dual anchors, it alleviates the unidirectional error accumulation plaguing existing methods. Experiments on ScanRefer and Nr3D benchmarks show that DACA, built entirely from open-source components, delivers outstanding zero-shot grounding performance, outperforming prior zero-shot approaches and exceeding most supervised baselines. Ablation studies validate the effectiveness of each module and offer new insights for spatiotemporal perception integrating semantic and geometric reasoning.

\clearpage

\section*{LLM Usage Statement}
In this work, large language models (LLMs) are only used as auxiliary tools for manuscript polishing and code formatting. They do not alter core arguments, algorithm logic, experimental configurations or conclusions, nor participate in study design, experimental implementation or result interpretation. The authors take full responsibility for all academic content and retain full control over the research.

{
	\small
	\bibliographystyle{ieeenat_fullname}
	\bibliography{cvpr2027_conference}
}

\AppendPDFPages{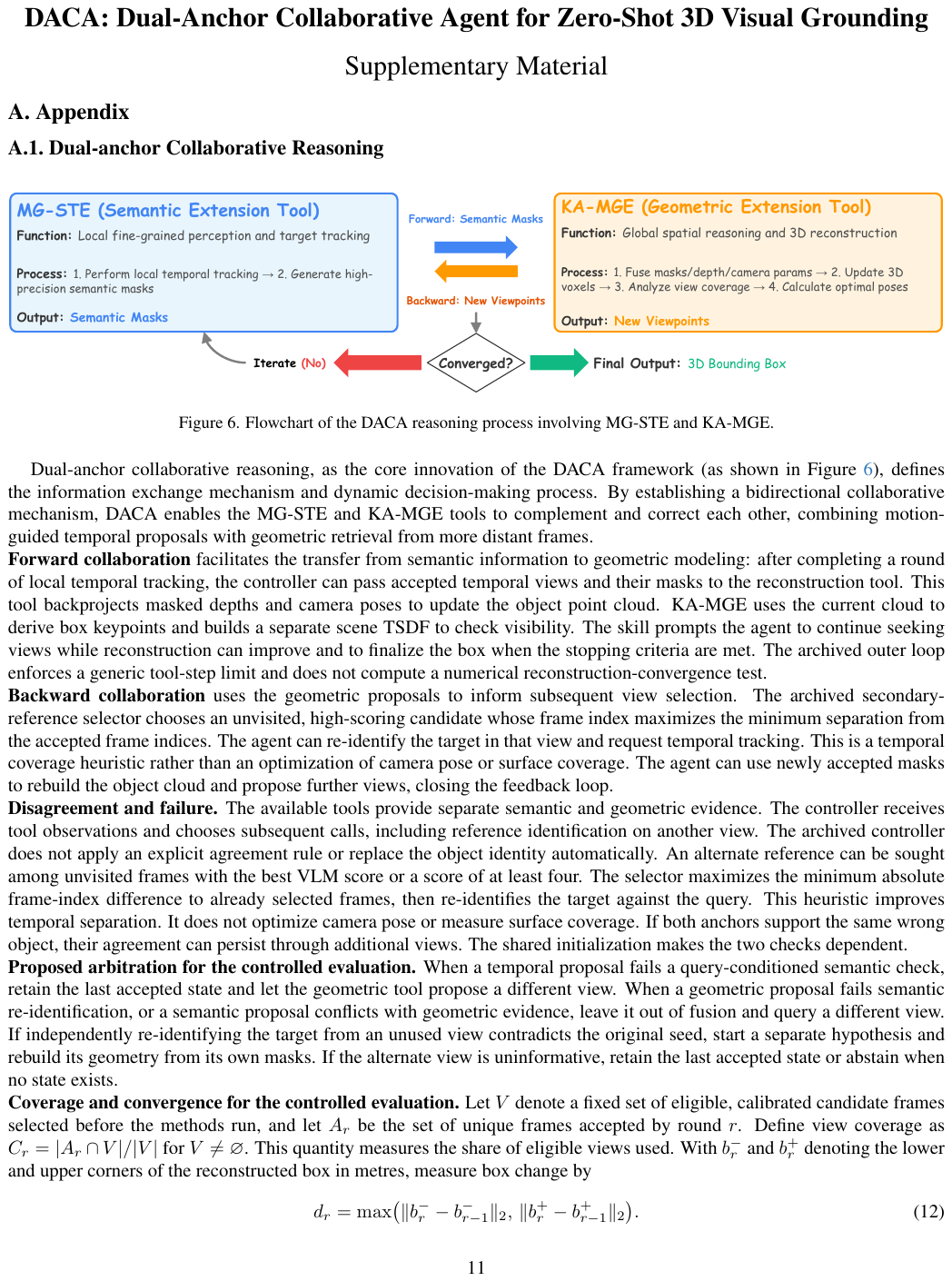}

\end{document}